\documentclass{ifacconf}

\usepackage{graphicx}      % include this line if your document contains figures
\usepackage{natbib}        % required for bibliography
\usepackage{booktabs}
\usepackage{graphicx}
\usepackage{subcaption}  % better than subfig
\usepackage{caption}
\usepackage[font=small]{caption}
\usepackage{amsmath}
\usepackage{amsfonts}
\usepackage{bm}
\usepackage{siunitx}

\usepackage{tikz}
\usepackage{multirow}
\usepackage{diagbox}
\begin{document}
\begin{frontmatter}

% \title{Style for IFAC Conferences \& Symposia: Use Title Case for
%   Paper Title\thanksref{footnoteinfo}} 
\title{Symbolic Learning of Interpretable Reduced-Order Models for Jumping Quadruped Robots}
% Title, preferably not more than 10 words.

% \thanks[footnoteinfo]{Sponsor and financial support acknowledgment
% goes here. Paper titles should be written in uppercase and lowercase
% letters, not all uppercase.}

\author[First]{Gioele Buriani$^{\dagger}$} 
\author[First]{Jingyue Liu$^{\dagger}$} 
\author[Second]{Maximilian St\"olzle}
\author[Third]{Cosimo Della Santina}
\author[Fourth]{Jiatao Ding}

\address[First]{Department of Cognitive Robotics, Delft University of Technology, Building 34, Mekelweg 2, 2628 CD Delft, Netherlands 
  (e-mail:  gioele.buriani@gmail.com,J.Liu-14@tudelft.nl.}

\address[Second]{Computer Science and Artificial Intelligence Laboratory (CSAIL), Massachusetts Institute of Technology, Cambridge, MA 02139 USA 
  (e-mail: M.W.Stolzle@tudelft.nl).}

\address[Third]{Institute of Robotics and Mechatronics, German Aerospace Center (DLR), 82234 Wessling, Germany 
  (e-mail: C.DellaSantina@tudelft.nl).}

\address[Fourth]{Department of Industrial Engineering, University of Trento (e-mail: jiatao.ding@unitn.it).}

\thanks{
This work is supported by the European Union’s Horizon Europe Program Project EMERGE - Grant Agreement No. 101070918.
}
\thanks{
$^{\dagger}$ These authors contributed equally to this work.
}

\begin{abstract}
Reduced-order models are central to motion planning and control of quadruped robots, yet existing templates are often hand-crafted for a specific locomotion modality. This motivates the need for automatic methods that extract task-specific, interpretable low-dimensional dynamics directly from data. We propose a methodology that combines a linear autoencoder with symbolic regression to derive such models. The linear autoencoder provides a consistent latent embedding for configurations, velocities, accelerations, and inputs, enabling the sparse identification of nonlinear dynamics (SINDy) to operate in a compact, physics-aligned space. A multi-phase, hybrid-aware training scheme ensures coherent latent coordinates across contact transitions. We focus our validation on quadruped jumping—a representative, challenging, yet contained scenario in which a principled template model is especially valuable. The resulting symbolic dynamics outperform the state-of-the-art handcrafted actuated spring-loaded inverted pendulum (aSLIP) baseline in simulation and hardware across multiple robots and jumping modalities.
\end{abstract}


\end{frontmatter}
%===============================================================================

\section{Introduction}
Legged robots are now pervasive systems, deployed in applications ranging from planetary exploration~\citep{arm2023scientific} to search-and-rescue operations~\citep{bai2018optional}. Yet, despite this growing presence, their dynamics remain remarkably difficult to model—especially when seeking a description expressive enough to capture essential behaviors and lightweight enough for real-time use.

Classic Lagrangian and Newton–Euler formulations provide full-order models of legged systems~\citep{bellicoso2017dynamic, di2018dynamic, yan2021whole, ding2024quadrupedal}, but these models remain too computationally heavy for online deployment~\citep{li2023real}. This has led to reduced-order templates such as spring-loaded inverted pendulum (SLIP)~\citep{blickhan1989spring} and single rigid body dynamics (SRBD)~\citep{hong2020real}, which capture dominant behaviors at far lower cost. Yet these templates lose accuracy in multi-contact transitions, rapid reconfiguration, and complex terrain~\citep{carpentier2018multicontact}. In practice, one must choose between high-dimensional models that are difficult to deploy and compact, task-specific heuristics. A principled way to automatically extract reduced-order models—accurate, compact, and interpretable—would significantly advance the analysis and control of hybrid locomotion dynamics. However, such a method is still missing.

In this paper, we turn to interpretable data-driven techniques to address this challenge. Data-driven learning has shown strong capabilities in modeling physical systems, with numerous works leveraging neural networks to capture complex dynamics~\citep{calandra2015learning, ogunmolu2016nonlinear, rajendra2020modeling, hashemi2023multibody}. Dimensionality-reduction methods such as autoencoders extract low-dimensional latent features~\citep{hinton2006reducing, gonzalez2018deep, refinetti2022dynamics}, after which multi-layer perceptrons (MLPs) or recurrent neural networks (RNNs) can model reduced-order dynamics—often gaining accuracy at the expense of interpretability. Physics-informed approaches, including physics-informed neural networks (PINNs)~\citep{raissi2019physics}, deep Lagrangian networks (DeLaN)~\citep{lutter2019deep, liu2024physics}, and coupled oscillator networks~\citep{stolzle2024input}, improve interpretability, data efficiency, and out-of-distribution performance. Still, none of these methods handle the hybrid, time-dependent structure inherent to locomotion. This limitation is particularly acute for floating-base dynamics, where contacts induce abrupt changes in inertial and potential forces—effects that current learning frameworks are not designed to capture.

Given this gap, no existing work provides a systematic method for learning reduced-order models of legged systems. We address this by combining linear autoencoders with symbolic regression, specifically the sparse identification of nonlinear dynamics (SINDy)~\citep{brunton2016discovering, pandey2024data}, extending prior autoencoder-based formulations~\citep{champion2019data} to quadruped dynamics, where such techniques remain unexplored.

The linear autoencoder is essential: its constant Jacobian—equal to its weight matrix—provides a unified linear mapping of configurations, velocities, accelerations, and inputs, enabling consistent and interpretable identification of reduced-order dynamics. Direct identification in the observation space is impractical due to scale disparities and dimensionality, while nonlinear autoencoders introduce state-dependent latent geometries and distort physical structure through separate nonlinear mappings of positions and velocities.

To obtain a representation valid across hybrid phases, we adopt a multi-phase training scheme: train the encoder in the contact phase (launch + landing), refine the decoder across all phases, and finally identify SINDy parameters with the autoencoder frozen. We validate the approach on jumping motions, a demanding form of dynamic locomotion that enables navigation in complex environments~\citep{nguyen2019optimized,atanassov2024curriculum,ding2023robust} yet remains challenging due to high accelerations and impact-driven dynamics.

In summary, this work contributes to legged robotics and physics-informed machine learning through:
\begin{itemize}
\item A physics-aligned linear latent embedding that provides consistent linear mappings of positions, velocities, accelerations, and inputs, enabling symbolic learning of reduced-order and interpretable jumping dynamics.
\item A sequential multi-phase training strategy that preserves coherent latent coordinates and symbolic models across the hybrid phases of a jump.
\item The first demonstration of learning reduced-order dynamics for legged locomotion, with evaluation across different quadruped robots and jumping modalities.
\end{itemize}

\section{Preliminary} \label{sec:Preliminary}

\subsection{Robot description}\label{sec:robot_des}
\begin{figure}[htb]
    \centering
    \includegraphics[width=0.8\linewidth]{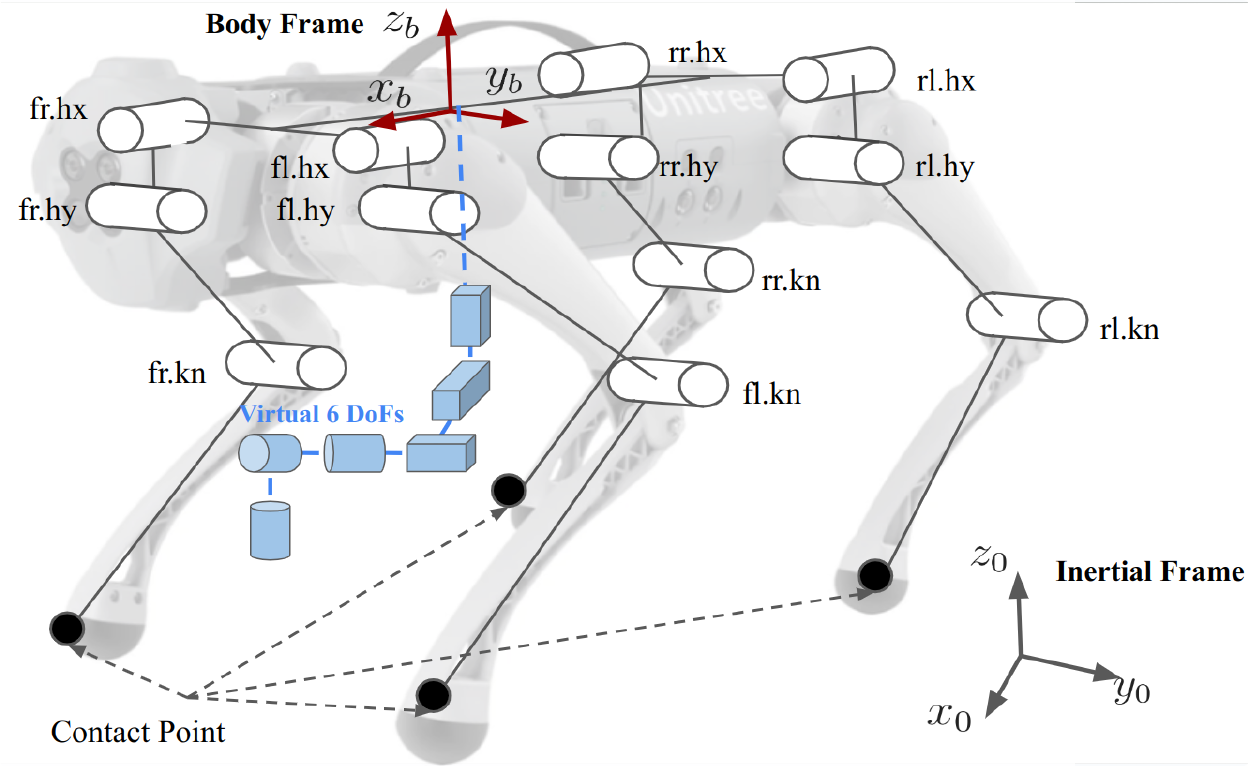}
    \vspace{-1.5mm}
    \caption{Quadruped robot with labeled joints and body frame.}
    \label{fig:float_base}
    \vspace{-1mm}
\end{figure}

Traditional modeling approaches represent the quadruped robot as a movable base (i.e., the body) with kinematic chains (i.e., the legs) attached, as shown in Fig.~\ref{fig:float_base}.

As illustrated in Fig.~\ref{fig:float_base}, the Unitree Go1, for example, exhibits $12$ degrees of freedom (DOFs) for the limbs alone, with $6$ underactuated DoFs for the floating base's position and orientation relative to an inertial frame~\citep{righetti2011inverse}. The configuration space of quadruped robots is typically parameterized as~\citep{mistry2010inverse}
\begin{equation}
    \mathbf{q} =
    \begin{bmatrix}
        \mathbf{q}_{\mathrm{j}}^\top &
        \mathbf{q}_{\mathrm{b}}^\top
    \end{bmatrix}^\top \in \mathbb{R}^{m +6},
\end{equation}
where $\mathbf{q}_{\mathrm{j}} \in \mathbb{R}^{m}$ contains the coordinates of the actuated joints, with $m$ representing the number of joints, and $\mathbf{q}_{\mathrm{b}} \in SE(3)$ captures the coordinates of the unactuated floating base, including Cartesian positions $(x, y, z) \in \mathbb{R}^3$ and Euler angles $(\phi, \theta, \psi) \in SO(3)$. The constrained dynamics follow
\begin{equation}
    \mathbf{M}(\mathbf{q}) \, \ddot{\mathbf{q}} + \mathbf{h}(\mathbf{q},\dot{\mathbf{q}}) = \mathbf{S}^\top \bm{\tau} + \mathbf{J}_{\tt C}^\top (\mathbf{q}) \, \bm{\lambda},
\end{equation}
where $\mathbf{M}(\mathbf{q}) \in \mathbb{R}^{(m + 6) \times (m + 6)}$ is the mass matrix, $\mathbf{h}(\mathbf{q},\dot{\mathbf{q}}) \in \mathbb{R}^{m + 6}$ captures centripetal, Coriolis, and gravitational forces, $\mathbf{S} = \begin{bmatrix} \bm{I}_{m \times m} & \bm{0}_{m \times 6} \end{bmatrix}$ is the selection matrix for actuated joints, $\bm{\tau} \in \mathbb{R}^{m}$ denotes actuation torques, $\mathbf{J}_{\tt C} \in \mathbb{R}^{k \times (m + 6)}$ is the Jacobian matrix for $k$ linearly independent constraints, and $\bm{\lambda} \in \mathbb{R}^k$ represents the constraint forces.

Consider a quadruped robot where ground reaction forces (GRFs) constitute the dominant constraint forces acting on the body. For each leg (modeled as an independent kinematic chain), the GRF $\mathbf{F}_k \in \mathbb{R}^3$ at the $k$-th foot can be computed from the joint torques $\bm{\tau}_k$ through the leg Jacobian transpose 
\begin{equation}
\mathbf{F}_k = \mathbf{J}_k^{-\top} \bm{\tau}_k,
\label{eq:grf_torque}
\end{equation}
where $\mathbf{J}_k \in \mathbb{R}^{3\times n_j}$ is the Jacobian of the $k$-th leg ($n_j = m/4$ being the number of joints per leg).

The net effect of GRFs on the robot's center of mass (CoM) dynamics can be described by the time derivatives of the linear and angular momentum of the CoM
\begin{equation}
\dot{\bm{\mathcal{P}}} = \sum_{k=1}^4 \mathbf{F}_k, \quad \dot{
\bm{L}
}= \sum_{k=1}^4 ((\mathbf{p}_k - \mathbf{b}) \times \mathbf{F}_k),
\label{eq.CoM_Linear_Angular}
\end{equation}
where $\bm{\mathcal{P}} \in \mathbb{R}^3$ and $\bm{L} \in \mathbb{R}^3$ are the linear and angular momentum, respectively. Here, \(\mathbf{p}_k \in \mathbb{R}^3 \) denotes the world-frame position of the $k$-th leg, and $\mathbf{b} \in \mathbb{R}^3$ is the CoM position in world coordinates.

The generalized wrench $\bm{\mathcal{F}} \in \mathbb{R}^6$ that acts on the body frame is
\begin{equation}
    \bm{\mathcal{F}} =
    \begin{bmatrix}
    \dot{\bm{\mathcal{P}}}^\top  & \dot{\bm{L}}^\top 
    \end{bmatrix}^\top  \in \mathbb{R}^6,
    \label{eq.CoM_force}
\end{equation}
which leads to the complete system input formulation
\begin{equation}
    \mathbf{u} = \begin{bmatrix}
    \bm{\tau}^\top & \bm{\mathcal{F}}^\top
    \end{bmatrix}^\top \in \mathbb{R}^{m + 6}.
\end{equation}

\subsection{Encoder–Decoder Representation Learning}\label{section:decoder}
The autoencoder, i.e., encoder--decoder architecture, enables one to obtain a compact but informative representation of high-dimensional state variables \citep{hinton2006reducing}. The encoder $E(\cdot)$ maps the original variable $\bm{x} \in \mathbb{R}^n$ into a low-dimensional latent vector $\bm{z} = E(\bm{x}) \in \mathbb{R}^k$ with $k \ll n$, effectively capturing the essential features that characterize the system's dynamics. The decoder $D(\cdot)$ reconstructs the high-dimensional signal from the latent representation, i.e., $\hat{\bm{x}} = D(\bm{z})$, ensuring that the latent space preserves the underlying structure of the original data. This framework not only reduces computational complexity but also provides a smooth and continuous latent manifold that is amenable to downstream tasks such as policy learning.

\subsection{SINDy}\label{section:sindy}
As a typical symbolic learning approach, SINDy~\citep{brunton2016discovering} enables us to identify symbolic representations of the dynamics by constructing a symbolic library of candidate functions $\Theta(\bm{\xi}, \dot{\bm{\xi}}, \bm{\nu})$, including polynomial and trigonometric functions of latent variables ($\bm{\xi}$), their derivatives ($\bm{\xi}^{(n)}$), and the transformed control inputs ($\bm{\nu}$). This library represents a comprehensive set of terms that might appear in the system's governing equations. The objective is to represent the dynamics of the latent variables as a sparse combination of these candidate functions. Mathematically, this is expressed as a sparse regression problem, with the latent dynamics expressed as
\begin{equation}\label{eq:encoder_mapping}
    \ddot{\bm{\xi}}_{\mathrm{pred}} = \Theta(\bm{\xi}, \dot{\bm{\xi}}, \bm{\nu}) \, \bm{\Xi},
\end{equation}
where $\bm{\Xi} \in \mathbb{R}^{p \times l}$ is a sparse matrix of coefficients to be determined, with $p$ being the number of the candidate functions in $\Theta(\bm{\xi}, \dot{\bm{\xi}}, \bm{\nu})$. The optimization problem is typically solved using sequential thresholding methods to ensure a sparse solution~\citep{brunton2016discovering, champion2019data}. Promotes sparsity by penalizing the $\ell_1$-norm of $\bm{\Xi}$, which encourages most entries in $\Xi$ to be zero. This sparsity-promoting optimization ensures that only a minimal set of candidate functions is selected, resulting in an interpretable model.

\section{Methodology}
\label{sec:Method}

\subsection{Proposed Learning Architecture}\label{sub:architecture}

\begin{figure}[htb]
        \centering
    \includegraphics[width=1\linewidth]{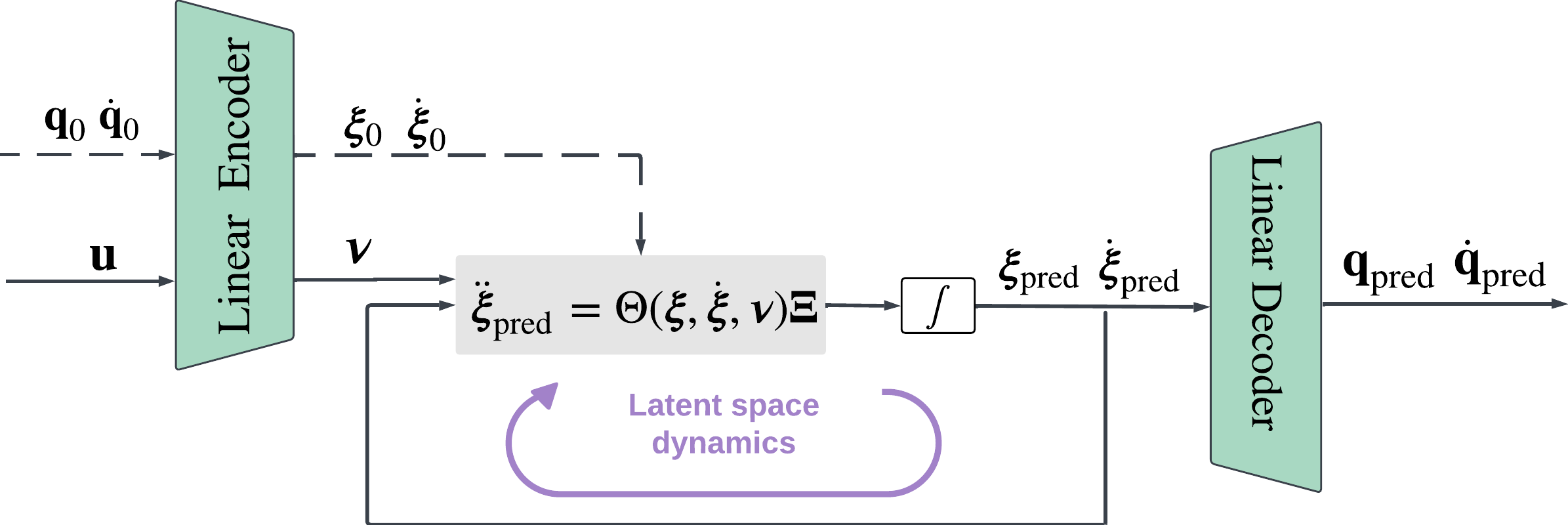}
    \vspace{-4.0mm}
    \caption{Methodology overview: Symbolic latent-space dynamics $\ddot{\bm{\xi}} = \bm{\Theta}(\bm{\xi}, \dot{\bm{\xi}}, \bm{\nu}) \, \bm{\Xi}$ are learned to predict the evolution of the full-state system from $(\mathbf{q}, \dot{\mathbf{q}})$. }
    % Here, $\mathbf{q}_0$ and $\dot{\mathbf{q}}_0$ represent the initial state.}
    \label{fig:whole_framework}
    \vspace{-1mm}
\end{figure}

We combine linear autoencoders with SINDy to model jumping dynamics in the latent space, achieving dimensionality reduction while preserving the essential dynamic characteristics, enabling efficient representation of complex jumping motions. Fig.~\ref{fig:whole_framework} provides an overview of the encoded latent space and the established dynamics loop. 

\begin{figure*}[htbp]
    \centering
    \begin{subfigure}[t]{0.5\textwidth}
    \centering
        \includegraphics[height=0.4\textwidth]{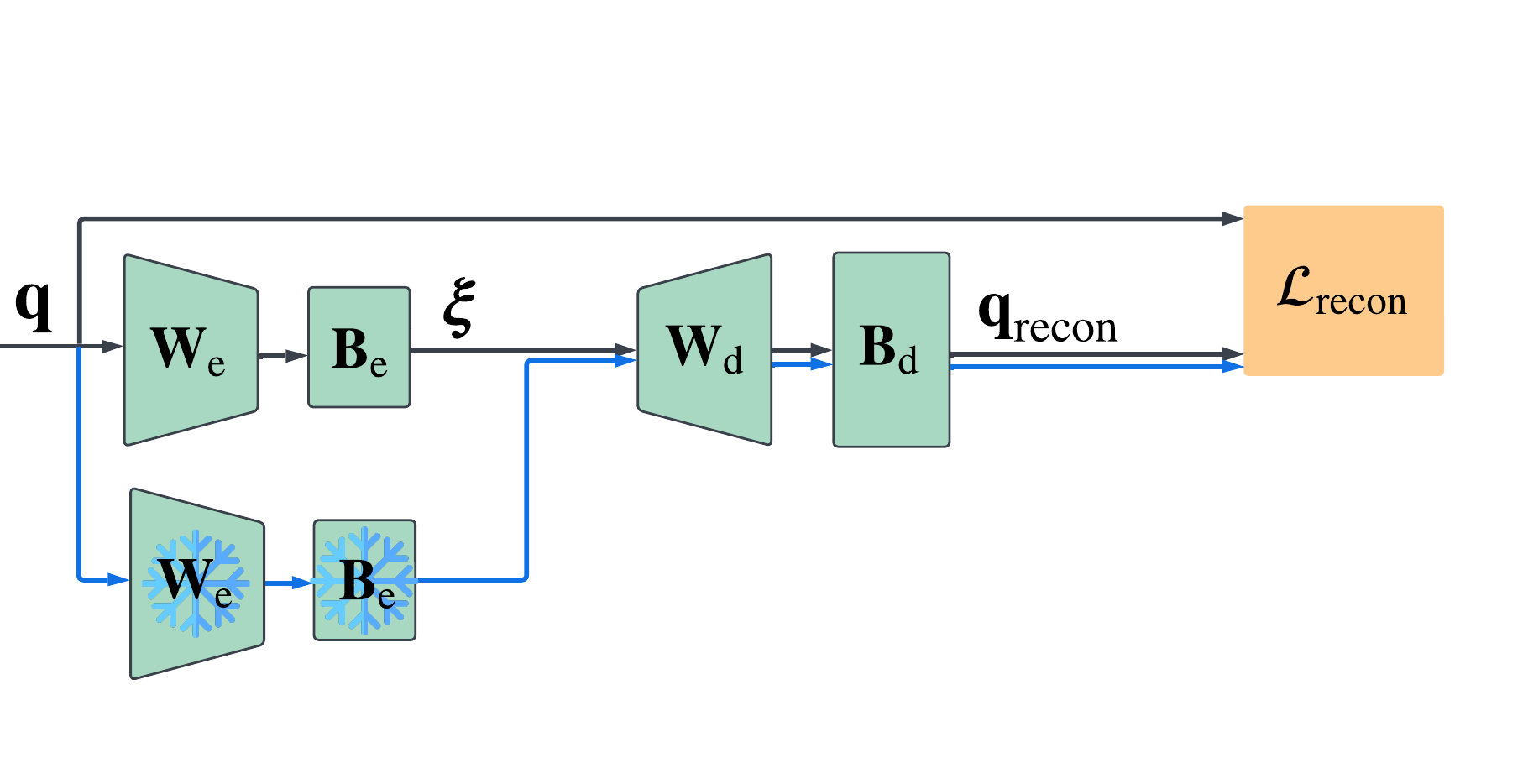}
        \caption{Training of encoder and decoder in a static setting}
        % \caption{Humanoid robot with labeled key joints and body frame.}
        \label{fig:step_1_2}
    \end{subfigure}%
    \hfill
    \begin{subfigure}[t]{0.5\textwidth}
        \centering
        \includegraphics[width=0.8\textwidth]{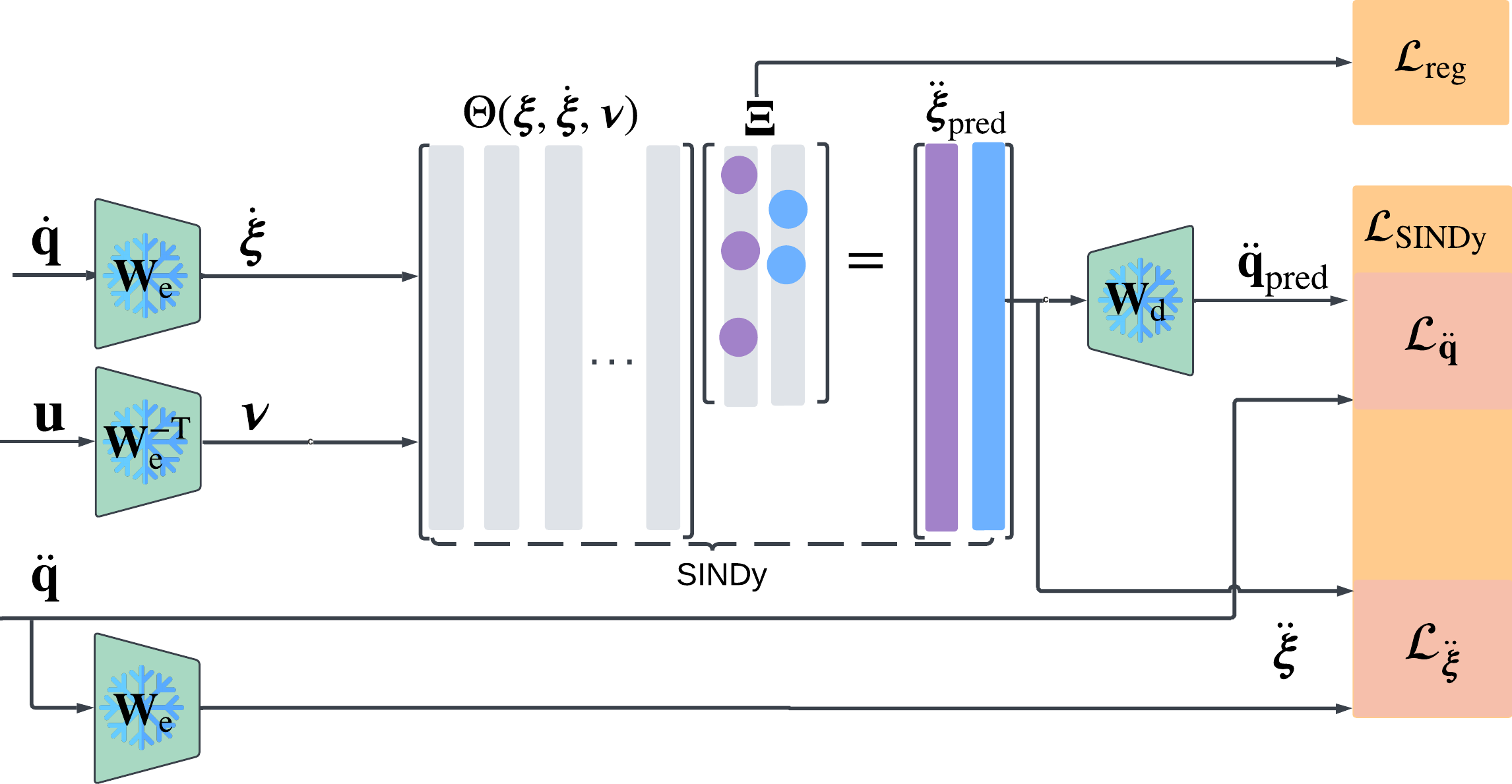}
        \caption{Training of symbolic latent dynamics in a dynamic setting}
        \label{fig:step3_SINDy}
    \end{subfigure}
    \vspace{-3mm}
    \caption{
    Flowchart of the pipeline. \textbf{(a)} The linear autoencoder is first trained on the contact phase dataset, minimizing the reconstruction loss, $\mathcal{L}_\mathrm{rec}$, depicted by the black line. Subsequently, by fixing the encoder, the decoder is fine-tuned on a dataset that includes all motion phases (see the the blue line). \textbf{(b)} In the third step, we train, for each motion phase separately, a symbolic dynamical model $\bm{\ddot{\xi}}_\text{pred}$ in latent space. Here, we employ the SINDy approach to regress a sparse matrix $\bm{\Xi}$ that selects and scales basis functions from the library $\bm{\Theta}(\bm{\xi}, \dot{\bm{\xi}}, \bm{\nu})$. 
    The loss components $\mathcal{L}_{\text{SINDy}}$ and $\mathcal{L}_{\text{reg}}$ ensure accurate dynamics prediction and sparsity of the coefficients, respectively. The SINDy loss $\mathcal{L}_{\text{SINDy}}$ is calculated based on the predicted reduced-order model accelerations, $\ddot{\bm{\xi}}_\mathrm{pred}$, and the corresponding decoded configuration-space accelerations, $\ddot{\mathbf{q}}_\mathrm{pred}$.
    % Importantly, we keep the autoencoder weights frozen, which preserves training stability and ensures that the reconstructions remain accurate during all motion phases.
    }
    \label{fig:training_algorithm}
\end{figure*}

Fig.~\ref{fig:training_algorithm} details the core components, including the encoder-decoder structure, latent space dynamics prediction, and real-space dynamics reconstruction. That is, we encode the initial state into a latent space, predict the dynamics in this latent space, and decode them back to the full-order system space. In particular, we ensure accurate reconstruction of system states and dynamical modeling through specialized loss components.

To maintain direct interpretability, we adopt the linear autoencoder to reduce the dimensionality of the state space while preserving a linear relationship with the original space. The autoencoder consists of an encoder $\left (\mathbf{W}_\mathrm{e}, \mathbf{B}_\mathrm{e} \right )$ and a decoder $\left (\mathbf{W}_\mathrm{d}, \mathbf{B}_\mathrm{d} \right )$. The encoder's weights and bias are $\mathbf{W}_\mathrm{e} \in \mathbb{R}^{l \times (m+6)}$ and $\mathbf{B}_\mathrm{e} \in \mathbb{R}^{l}$, while the decoder's weights and bias are $\mathbf{W}_\mathrm{d} \in \mathbb{R}^{(m+6) \times l}$ and $\mathbf{B}_\mathrm{d} \in \mathbb{R}^{m+6}$, where $l$ is the dimension of the desired latent space. 

With this encoder, we map the robot's configuration $\mathbf{q}$ and its derivatives $\mathbf{q}^{(n)}$ into a lower-dimensional latent space using the linear transformation
\begin{equation}
    \bm{\xi}^{(n)} =  \mathbf{W}_\mathrm{e} \, \mathbf{q}^{(n)} + \delta_{n0} \, \mathbf{B}_\mathrm{e}, \quad n = 0, 1, 2,
\end{equation}
where $\bm{\xi} \in \mathbb{R}^{l}$ represents the latent representation, $\bm{\xi}^{(n)}$ denotes the $n$-th derivative of $\bm{\xi}$, and $\delta_{n0}$ is the Kronecker delta, equal to $1$ when $n = 0$ and $0$ otherwise. 
The system's actuation is mapped into latent space as
\begin{equation}
\begin{array}{rl}
     \bm{\nu} & = \mathbf{W}_\mathrm{e}^{-\top} \mathbf{u} \in \mathbb{R}^l.
\end{array}
\end{equation}

After establishing the latent space mapping, we employ the SINDy algorithm to identify symbolic representations of the latent dynamics (see Fig.~\ref{fig:step3_SINDy}). Particularly, we follow Eq.~\eqref{eq:encoder_mapping} to predict the accelerations. Subsequently, the predicted accelerations $\ddot{\bm{\xi}}_{\mathrm{pred}}$ are transformed back to the original state space by the linear decoder
\begin{equation}
\ddot{\mathbf{q}}_{\mathrm{pred}} = \mathbf{W}_\mathrm{d} \, \ddot{\bm{\xi}}_{\mathrm{pred}}.
\end{equation}
For training the encoder, decoder, and reduced-order symbolic dynamics, we consider the following loss components:
\begin{itemize}
    \item $\mathcal{L}_{\mathrm{recon}} = \left\| \mathbf{q} - \mathbf{q}_{\mathrm{recon}} \right\|_2^2$ ensures accurate reconstruction of the original state. The reconstructed configuration $\mathbf{q}_{\mathrm{recon}}$ is given by
    \begin{equation}
        \mathbf{q}_{\mathrm{recon}} = \mathbf{W}_\mathrm{d} \, (\mathbf{W}_\mathrm{e}\mathbf{q} + \mathbf{B}_\mathrm{e}) + \mathbf{B}_\mathrm{d}. 
        \label{eq.reconstruct}
    \end{equation}
    \item $\mathcal{L}_{\mathrm{SINDy}} = \| \ddot{\bm{\xi}} - \ddot{\bm{\xi}}_{\mathrm{pred}} \|_2^2 + \left\| \ddot{\mathbf{q}} - \ddot{\mathbf{q}}_{\mathrm{pred}} \right\|_2^2 $ captures the prediction loss, ensuring that the learned dynamics in the latent space and the reconstructed space are accurate.
    \item $\mathcal{L}_{\mathrm{reg}} = \| \bm{\Xi} \|_1$ promotes sparsity in SINDy coefficients.
\end{itemize}

\subsection{Multi-Phase Training Strategy}\label{subsec.training}

Jumping motion poses unique modeling challenges due to its hybrid dynamical nature. This study considers two types of jumping: the pronking and the froggy jumps.

In the pronking jump, the robot launches and lands with full limb contact simultaneously. This motion consists of two dominant phases: (1) a contact phase, where all legs remain grounded during takeoff and landing, and (2) a flight phase, where the robot is entirely airborne. Conversely, the froggy jump consists of an initial phase where all legs are grounded, followed by a transition phase where only the rear legs remain in contact with the ground, and then progressing to a flight phase. 

To effectively model the hybrid dynamics, we propose a structured training methodology (Fig.~\ref{fig:training_algorithm}) that combines transfer learning principles~\citep{torrey2010transfer} with phase-specific optimization. First, we train the encoder and decoder statically on reconstructing the configurations of the full contact phase by minimizing the $\mathcal{L}_\mathrm{recon}$ loss. Then, the encoder weights are frozen, and the decoder is trained on all motion phases, including the flight phase, while still minimizing $\mathcal{L}_\mathrm{recon}$. Finally, with the encoder and decoder weights fixed, we train the symbolic dynamics for each motion phase (contact, partial contact, and flight) by minimizing the $\mathcal{L}_\mathrm{SINDy}$ and $\mathcal{L}_\mathrm{reg}$ loss terms. In this way, the final model is physically coherent and can accurately reconstruct the robot's motion across different phases. %\red{$?$}By leveraging transfer learning, we create robust models that effectively capture hybrid dynamics, enhance interpretability, and seamlessly handle complex, phase-dependent behaviors.

\subsection{Data Collection and Preprocessing}
\label{subsec:data_coll}
Datasets were collected from both simulated and real-world environments. Simulated data were generated using the PyBullet engine~\citep{coumans2016pybullet} with Unitree Go1 and A1 robots. Real-world data were gathered from an actual Go1 robot. %to validate the simulation results, ensure practical applicability, and benchmark the algorithm's performance on real-world, noisy data.

\begin{figure*}[htb]
        \centering
    \subfloat[Simulated Go1]{
        \includegraphics[width=0.32\linewidth]{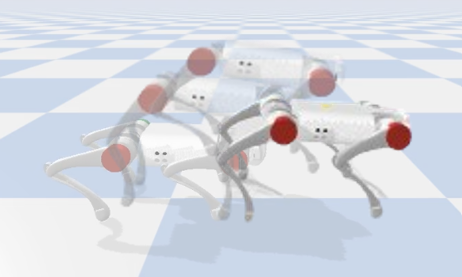}
        \label{fig:go1_sim}
    }
    \subfloat[Simulated A1 ]{
        \includegraphics[width=0.32\linewidth]{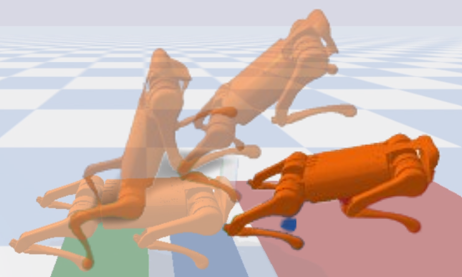}
        \label{fig:a1_sim}
    }
    \subfloat[Real Go1]{
        \includegraphics[width=0.32\linewidth]{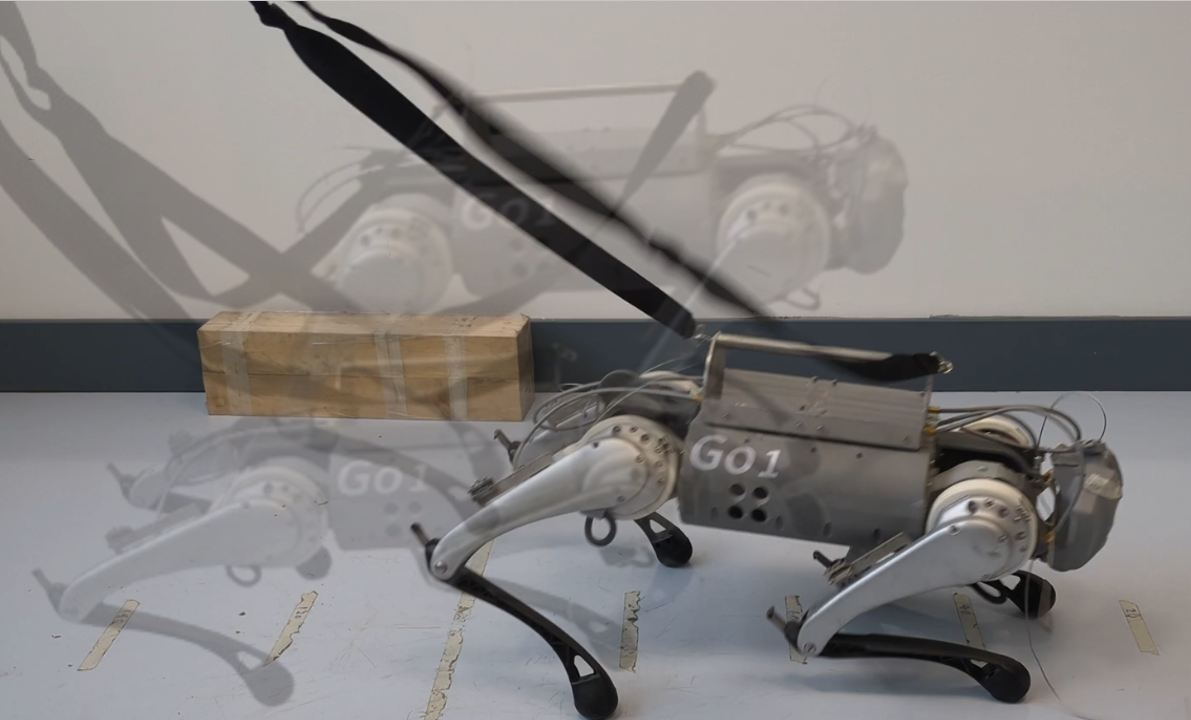}
        \label{fig:real_jump}
    }
    \vspace{-3mm}
    \caption{Sequential images of quadruped jumps:  \textbf{(a)} A simulated Go1 robot performs a pronking jump. \textbf{(b)} A simulated A1 executes a froggy jump. \textbf{(c)} A real Go1 performs a pronking jump.}
    \label{fig:subfigures}
\end{figure*}

Simulation data for Go1 captured forward pronking motions (see Fig.~\ref{fig:go1_sim}), achieved with model predictive control (MPC)~\citep{ding2024robust}. Data from A1 for froggy jumping (see Fig.~\ref{fig:a1_sim}) were collected using a Cartesian PD controller~\citep{nguyen2022contact}. The dataset includes joint angles and velocities ($\mathbf{q}_{\mathrm{j}}, \dot{\mathbf{q}}_{\mathrm{j}} \in \mathbb{R}^{12}$), body position and velocity ($\mathbf{q}_{\mathrm{b}}, \dot{\mathbf{q}}_{\mathrm{b}} \in \mathbb{R}^{6}$), input torques ($\bm{\tau} \in \mathbb{R}^{12}$), and the contact state ($\mathbf{c} \in \mathbb{R}^{4}$) for each timestep. 

The Go1 simulated data consists of 156 jumps (120 for training, 30 for validation, and 6 for testing), with each jump having 800 points: 300 for launching, 300 for landing, and 200 for flight. The A1 dataset contains 150 jumps (100 for training, 40 for validation, and 10 for testing), with each jump comprising 1200 points: 400 for launch, 400 for landing, and 400 for flight. To enhance robustness, a variant of the datasets with added noise was considered, as described in the work from \citet{panichi2025fly}.
% \begin{itemize}
%     \item Gaussian noise on the initial configuration 
%     \item Gaussian noise on the state readings
%     \item External disturbance force applied to the robot's CoM 
% \end{itemize}

Real-world data were collected from a Go1 robot, as depicted in Fig.~\ref{fig:real_jump}. The quadruped was operated using the same control algorithm as in the simulation, with data collection performed similarly using the official Go1 SDK. This dataset comprises 15 jumps, distributed as 12 for training, 2 for validation, and 1 for testing, maintaining the same data point distribution as the Go1 simulated data.

The collected data are preprocessed to compute necessary derivatives and forces. The state's acceleration $\ddot{\mathbf{q}}$ is calculated by numerically differentiating the velocity data $\dot{\mathbf{q}}$. The force on the CoM, $\bm{\mathcal{F}} \in \mathbb{R}^6$, is calculated based on Eqs.~\eqref{eq.CoM_Linear_Angular} and ~\eqref{eq.CoM_force}, forming the final system input, $\mathbf{u}$. 

\subsection{Evaluation Procedure}
\label{subsec:testing}
To validate the model $\Theta$, the initial state $(\mathbf{q}_0, \dot{\mathbf{q}}_0)$ and inputs $\mathbf{u}$ are transformed by the trained encoder and the input transformation equation to the latent initial state $(\bm{\xi}_0, \dot{\bm{\xi}}_0)$ and $\bm{\nu}$, respectively. The model then predicts latent accelerations $\ddot{\bm{\xi}}_{\text{pred}}$, which are integrated using the \emph{lsoda} method~\citep{hindmarsh2005lsodar} to obtain predicted latent space state ${\bm{\xi}}_{\text{pred}}$ and $\dot{\bm{\xi}}_{\text{pred}}$ with a step size of \SI{500}{Hz}. This process is repeated for each time step. The predicted latent configuration $\bm{\xi}_{\text{pred}}$ is then decoded into a original configuration space by $\mathbf{q}_{\text{pred}} = \mathbf{W}_\mathrm{d} \, \bm{\xi}_{\text{pred}} + \mathbf{B}_\mathrm{d}$ for direct comparison with actual jump data. 

An alternative evaluation approach involves simulating the learned system for shorter intervals (e.g., \SI{0.1}{s} -- 50 steps) with periodic resets to the original latent state $(\bm{\xi}, \dot{\bm{\xi}})$. This strategy decreases error accumulation over time, showcasing the precision of the model for medium-horizon forecasts, which could be used in planning and control frameworks such as MPC.

\section{Results}
\label{sec:Evaulation}
This section evaluates the
approach. We explored several latent-space dimensionalities to assess the trade-off between model compactness and accuracy. Two representative configurations are reported in Table~\ref{tab:hyperparameters}.
    \begin{table}[t]
    \centering
    \caption{Hyperparameters of the models. 
    The number of candidate functions is computed as  $\sum_{k=0}^{p} (2l+k-1) +2(2l)+l$ for latent dimension $l$ and polynomial order $p$.}
    \label{tab:hyperparameters}
    \vspace{-3mm}
    \begin{tabular}{lccccc}
        \toprule
        \textbf{Parameter} & \textbf{Symbol / Value} & \textbf{Go1} & \textbf{A1} &\\
        \midrule
        Latent dimension & $l$ & 2 & 4 \\
        Polynomial order & $p$ & 1 & 1 \\
        \# Candidates & $|\Theta|$ & 15 & 29 \\
        Sparsity threshold & $\lambda$ & \multicolumn{2}{c}{0.01}\\
        Learning rate & $\eta$ & \multicolumn{2}{c}{$1\times10^{-3}$}\\
        Batch size & $B$ & \multicolumn{2}{c}{100}  \\
        Epochs & $N_{\text{epoch}}$ & \multicolumn{2}{c}{501} \\
        % Solver & --- & \multicolumn{2}{c}{RK2} \\
        \bottomrule
        \end{tabular}
    \end{table}
% \subsection{Baseline: actuated SLIP Model}
% \label{subsec:reduced_order_model}
% We compare the learned model with a reduced-order model, e.g., the actuated SLIP (aSLIP)~\citep{ding2024robust} model, which is used for running, hopping, and jumping motion. The aSLIP model, extending the basic SLIP model by incorporating actuators, allows for enhanced control over legged jumping, as depicted in Fig.~\ref{fig:aslip}. 

% \begin{figure}[htb]
%     \centering
%     \includegraphics[width=0.35\linewidth]{Images/ASLIP.pdf}
%     \caption{ The aSLIP model for jumping motion of legged robots. The actuation (denoted as $m\mathbf{u}$) is explicitly captured to enhance controllability.}
%     \label{fig:aslip}
% \end{figure}
% The CoM acceleration $\mathbf{\ddot{b}}$ is 
% \begin{equation}
%     \begin{array}{r l}
%         \text{In flight: } &  \mathbf{\ddot{b}} =  \bm{g}\\
%         \text{In Contact: }&  \mathbf{\ddot{b}} =   \frac{k_\mathrm{s} |\bm{l}_0 - \bm{l}|\hat{\bm{l}}}{m} + \bm{g} + \bm{u}
%     \end{array}
% \end{equation}
% where $k_\mathrm{s} \in \mathbb{R}^+$ is the spring constant, $m \in \mathbb{R}^+$ is the body mass, $\bm{l}  = \mathbf{b} - \mathbf{p}_k$ is the leg vector, $\hat{\bm{l}}$ is the unit vector
% along the leg retraction direction,  $\bm{l}_0$ is the rest length, and $\bm{g} = [0, 0, -g]^\top$ with $g$ being the vertical gravitational constant.
% $\bm{u} \in \mathbb{R}^3$ are the driving forces generated by joint motors, which can be seen as the accumulated GRFs on four legs, as calculated using \eqref{eq:grf_torque}.

\subsection{Go1 Simulation Results}
\label{subsec:go1_simulation}
\subsubsection{Learning Go1 pronking}
\begin{figure}
    \centering
    \includegraphics[width=1\linewidth]{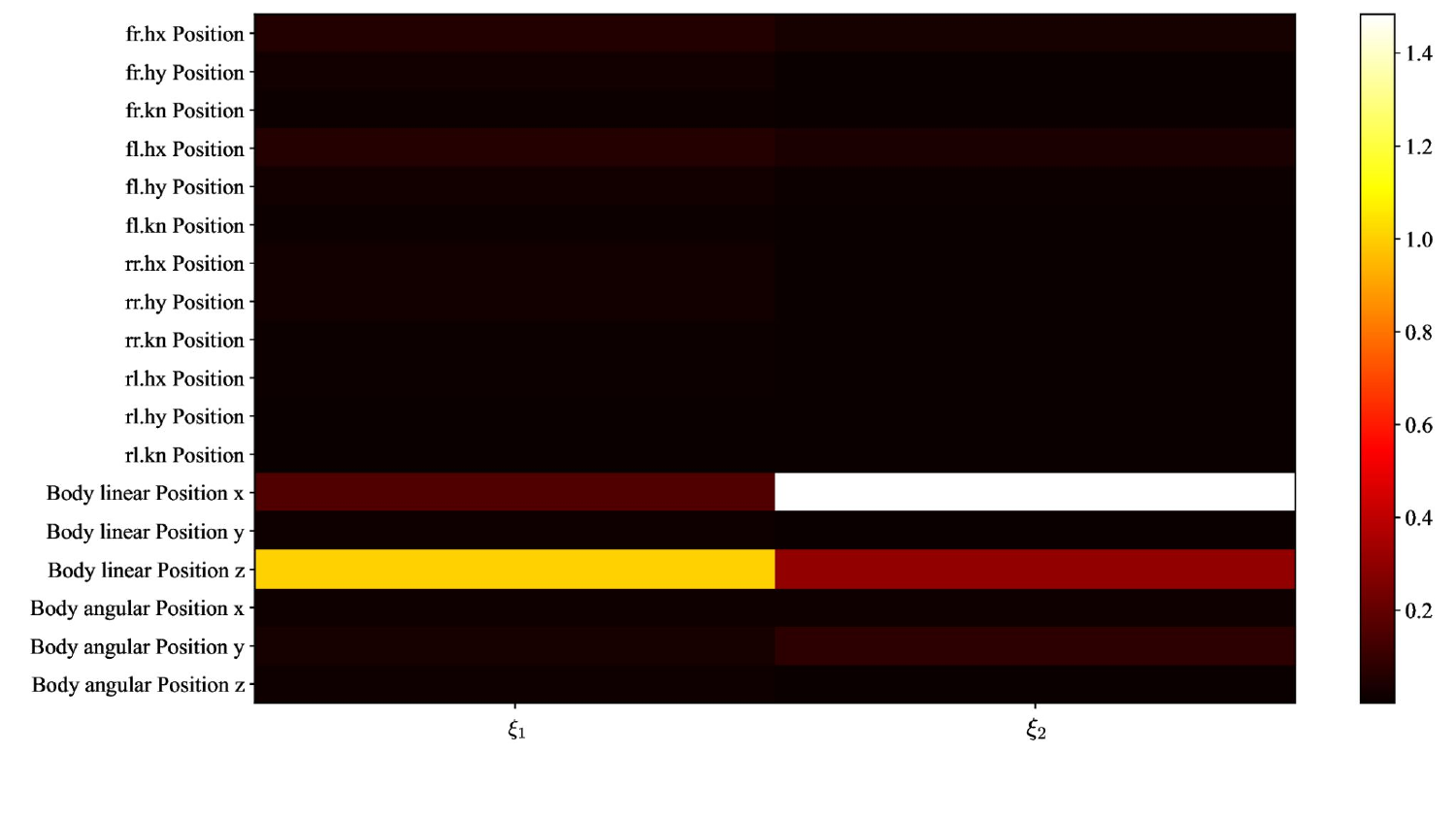}
    \vspace{-10mm}
    \caption{Heatmap of the encoder weight matrix $\mathbf{W}_\mathrm{e}^\mathrm{T}$ (Go1 simulation, two latent dimensions). Brighter colors indicate larger weights; rows are 18 configuration dimensions and columns are the two latent dimensions. }
    \label{fig:2d_enc}
\end{figure}

We initially explored a two-dimensional latent model for the pronking.
Fig.~\ref{fig:2d_enc} visualizes the transposed weight matrix $\mathbf{W}_\mathrm{e}^\mathrm{T}$ of the trained encoder, indicating that the latent dimensions primarily correspond to the CoM positions in the $x$ and $z$ coordinates.

The learned equations of motion in the latent space for the contact and flight phases, respectively, are
\begin{equation}
    \begin{cases}
        \ddot{\xi}_1\!\! =\!\! 0.36 \! - \!0.16 \, \dot{\xi}_1 \! - \! \!0.91 \, \dot{\xi}_2 \! - \! 0.75\sin (\dot{\xi}_1) \! + \! 0.11 \, \nu_1 \! - \! 0.05 \, \nu_2, \\
        \ddot{\xi}_2\!\! =\!\! -0.16\! -\! 0.23 \, \dot{\xi}_1 \!- \!0.75 \, \dot{\xi}_2\! - \!0.96 \, \sin (\dot{\xi}_2) \! + \!0.14 \, \nu_2,
    \end{cases}
\end{equation}
and 
\begin{equation}
    \begin{cases}
        \ddot{\xi}_1 = 0.29 - 1.22 \, \dot{\xi}_1 - 1.00 \sin (\dot{\xi}_1)  + 51.05 \, \nu_1, \\
        \ddot{\xi}_2 = 0.42 \, \dot{\xi}_1 + 0.52 \sin (\dot{\xi}_1).
    \end{cases}
\end{equation}
The latent coordinates do not correspond to explicit physical variables, but the learned symbolic equations display structures that are consistent with key principles of quadruped jumping. The constant terms behave like gravity-related biases, the linear velocity terms act as effective damping, %capturing energy dissipation, 
and the input-dependent components reflect how actuation influences the motion. The trigonometric functions of latent velocities serve as smooth nonlinear shaping terms that encode phase-dependent effects arising from posture changes, contact transitions, and limb coordination. These nonlinear components naturally emerge when compressing high-dimensional multibody dynamics into a low-dimensional latent representation.
\begin{figure}[htb]
    \centering
    \subfloat[Joint angles (Joint notations are detailed in Fig.~\ref{fig:float_base})]{
    \vspace{-2mm}
        \includegraphics[width=0.49\textwidth, trim=0 400 0 350, clip]{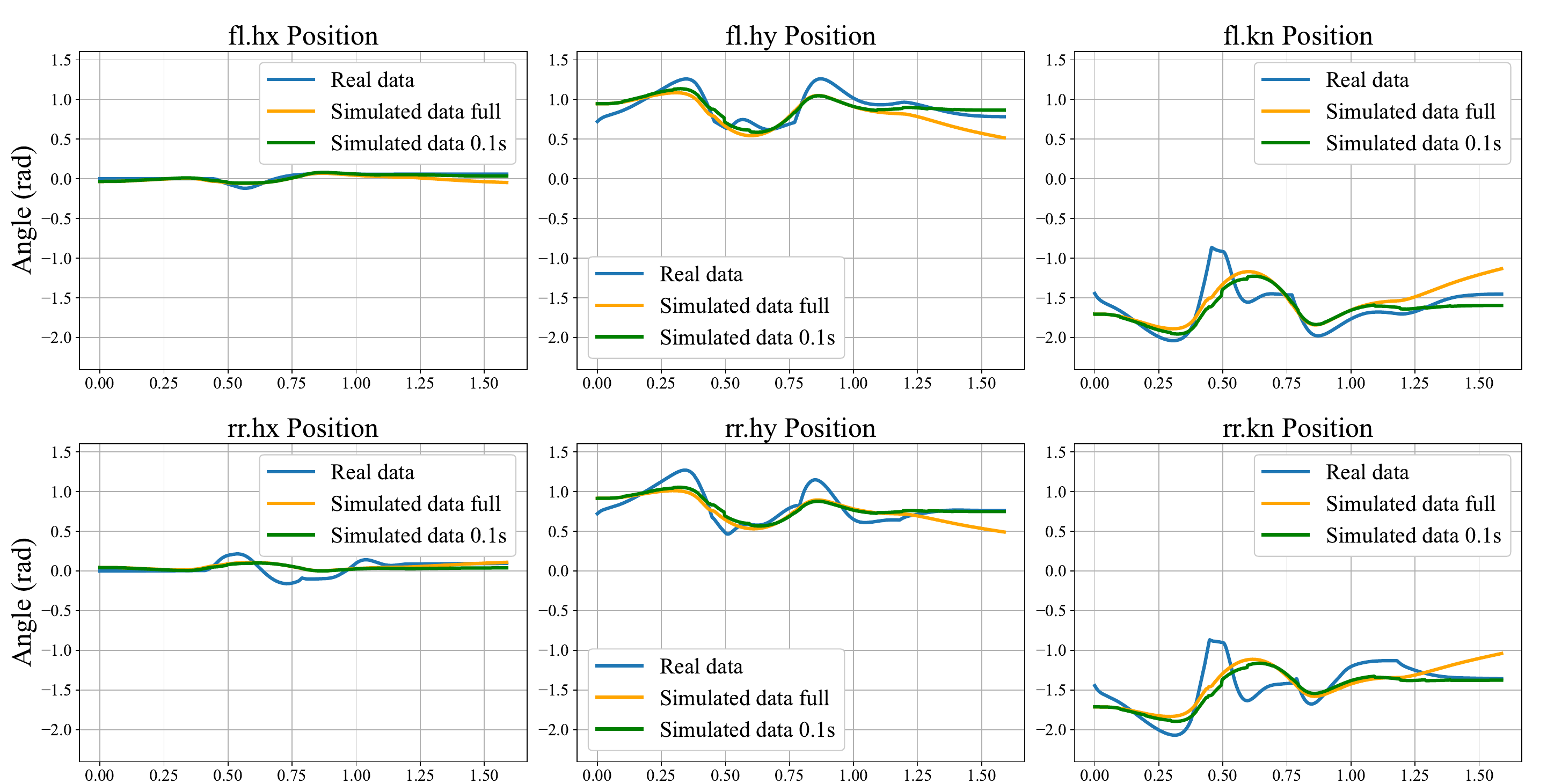}
    }\\

    \subfloat[CoM positions and orientations]{
    \vspace{-2mm}
        \includegraphics[width=0.49\textwidth]{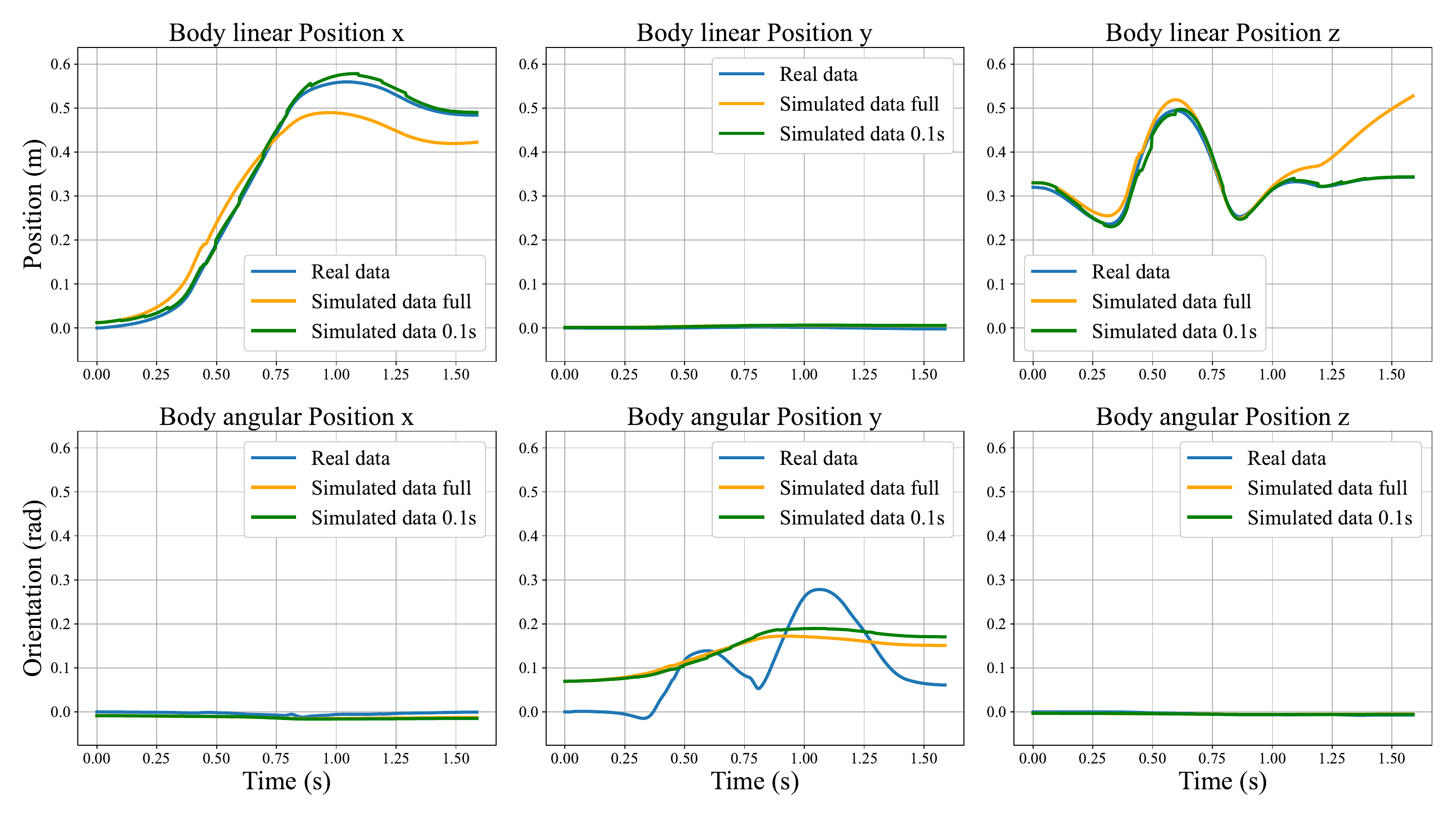}
    }
    \vspace{-2mm}
    \caption{Evaluation of the trained model (with two latent dimensions) for simulated Go1 pronking. Considering the symmetry between right and left side, we only report the joints angle profiles in front left (fl) and rear right (rr) leg.}
    \label{fig:2d_plots}
\end{figure}
% \begin{table}[b]
%   \centering
%   \renewcommand{\arraystretch}{1.25}
%   \caption{Prediction performance of the trained model with the simulation Go1 data.}
%   \label{tab:result_simulate_go1}  
%   \resizebox{0.49\textwidth}{!}{
%   \begin{tabular}{|c|c|c|c|c|c|c|}
%     \hline  
%     \diagbox[width=7em,dir=NW]{Pre.}{Joint} & fl hx & fl hy & fl hz & rr hx & rr hy & rr hz \\ 
%     \hline
%     Full& 1.3e-2$\pm$1.5e-2 & -8.0e-2$\pm$0.11 & 7.5e-2$\pm$0.18 & 2.3e-2$\pm$7.6e-2& -5.4e-2$\pm$0.13& 3.7e-2$\pm$0.22 \\ 
%     \hline
%     0.1s& 5.0e-3$\pm$7.6e-3 & -4.1e-3$\pm$9.4e-2 & -2.1e-2$\pm$0.17 & 4.4e-3$\pm$8.5e-2& -5.4e-3$\pm$0.11& -2.5e-2$\pm$0.21\\ 
%     \hline
%     \diagbox[width=7em,dir=NW]{Pre.}{CoM} & x & y & z & roll & pitch & yaw \\ 
%     \hline
%     Full& -2.2e-2$\pm$5.0e-2 & 4.4e-3$\pm$1.3e-3 & 4.1e-2$\pm$5.0e-2 & -9.8e-3$\pm$1.8e-3& 3.0e-2$\pm$6.4e-2& -1.5e-4$\pm$8.1e-4 \\ 
%     \hline
%     0.1s& 8.9e-3$\pm$4.6e-3 & 4.3e-3$\pm$1.3e-3 & 5.4e-4$\pm$9.4e-3 & -9.3e-3$\pm$2.6e-3& 3.8e-2$\pm$6.1e-2& -1.5e-4$\pm$8.1e-4\\ 
%     \hline    
%   \end{tabular}
%   }
% \end{table}

\begin{table*}[htb]
  \centering
  \renewcommand{\arraystretch}{1.25}
  \caption{Prediction performance of the trained model on simulated and real datasets. Values denote mean absolute error $\pm$ standard deviation for joint angles and CoM pose components.}
  \label{tab:result}  
  \resizebox{0.95\textwidth}{!}{
  \begin{tabular}{cccccccccccccc}
    \toprule  
      \multirow{2}{*}{Dataset} &  \multirow{2}{*}{Pre.} & \multicolumn{6}{c}{Joint angles} & \multicolumn{3}{c}{CoM position} & \multicolumn{3}{c}{CoM orientation} \\
    \cmidrule(lr){3-8} \cmidrule(lr){9-11} \cmidrule(lr){12-14} &  
            & fl hx & fl hy & fl hz & rr hx & rr hy & rr hz
            & x & y & z
            & roll & pitch & yaw \\
    \midrule
    \multirow{2}{*}{SIM Go1} & Full&  1.3e-2$\pm$1.5e-2 & -8.0e-2$\pm$0.11 & 7.5e-2$\pm$0.18 & 2.3e-2$\pm$7.6e-2& -5.4e-2$\pm$0.13& 3.7e-2$\pm$0.22 & -2.2e-2$\pm$5.0e-2 & 4.4e-3$\pm$1.3e-3 & 4.1e-2$\pm$5.0e-2 & -9.8e-3$\pm$1.8e-3& 3.0e-2$\pm$6.4e-2& -1.5e-4$\pm$8.1e-4\\ 
    % \hline
    & 0.1s & 5.0e-3$\pm$7.6e-3 & -4.1e-3$\pm$9.4e-2 & -2.1e-2$\pm$0.17 & 4.4e-3$\pm$8.5e-2& -5.4e-3$\pm$0.11& -2.5e-2$\pm$0.21 & 8.9e-3$\pm$4.6e-3 & 4.3e-3$\pm$1.3e-3 & 5.4e-4$\pm$9.4e-3 & -9.3e-3$\pm$2.6e-3& 3.8e-2$\pm$6.1e-2& -1.5e-4$\pm$8.1e-4\\ 
    \midrule
    \multirow{2}{*}{SIM A1} & Full& 1.0e-10$\pm$1.1e-10 & 0.19$\pm$0.16 & -0.30$\pm$0.34 & 1.0e-10$\pm$1.1e-10 & 0.26$\pm$0.20& -0.32$\pm$0.36 & -3.5e-2$\pm$5.1e-2 & -8.0e-3$\pm$1.5e-3 & -6.1e-2$\pm$5.8e-2 & 2.4e-3$\pm$4.7e-3& 0.10$\pm$8.3e-2& -8.7e-4$\pm$3.6e-3 \\ 
    % \hline
    & 0.1s & 1.0e-10$\pm$2.6e-10 & 2.8e-3$\pm$3.9e-2 & -3.0e-2$\pm$0.16 & 1.0e-10$\pm$2.6e-10 & 1.4e-2$\pm$5.2e-2& -3.9e-2$\pm$0.23 & -7.9e-3$\pm$1.2e-3 & -8.0e-3$\pm$1.5e-3 & -2.9e-3$\pm$2.0e-3 & 2.4e-3$\pm$4.7e-3& 9.8e-3$\pm$5.6e-2& -8.7e-4$\pm$3.6e-3\\ 
    \midrule
     \multirow{2}{*}{Real Go1} & Full& -3.7e-2$\pm$2.6e-2 & 0.22$\pm$0.15 & -0.35$\pm$0.39 & -4.2e-2$\pm$4.7e-2& 0.34$\pm$0.18& -0.65$\pm$0.49 & 7.5e-2$\pm$7.0e-2 & -9.6e-4$\pm$9.6e-3 & -8.1e-2$\pm$7.7e-2 & 3.3e-2$\pm$2.2e-2& -0.14$\pm$9.2e-2& 0.1$\pm$7.8e-2\\ 
    % \hline
    & 0.1s& -4.6e-3$\pm$2.3e-2 & 1.7e-2$\pm$8.3e-2 & 3.8e-2$\pm$0.19 &-7.5e-2$\pm$5.1e-2& 0.12$\pm$5.6e-2& -0.21$\pm$0.17 & 1.2e-3$\pm$3.1e-3 & -7.1e-3$\pm$5.4e-3 & 1.2e-3$\pm$3.4e-3 & 2.9e-2$\pm$1.9e-2& -0.10$\pm$5.9e-2& 0.12$\pm$8.9e-2\\ 
    \bottomrule
  \end{tabular}
  }
\end{table*}

\begin{figure}
    \centering
    \includegraphics[width=1\linewidth]{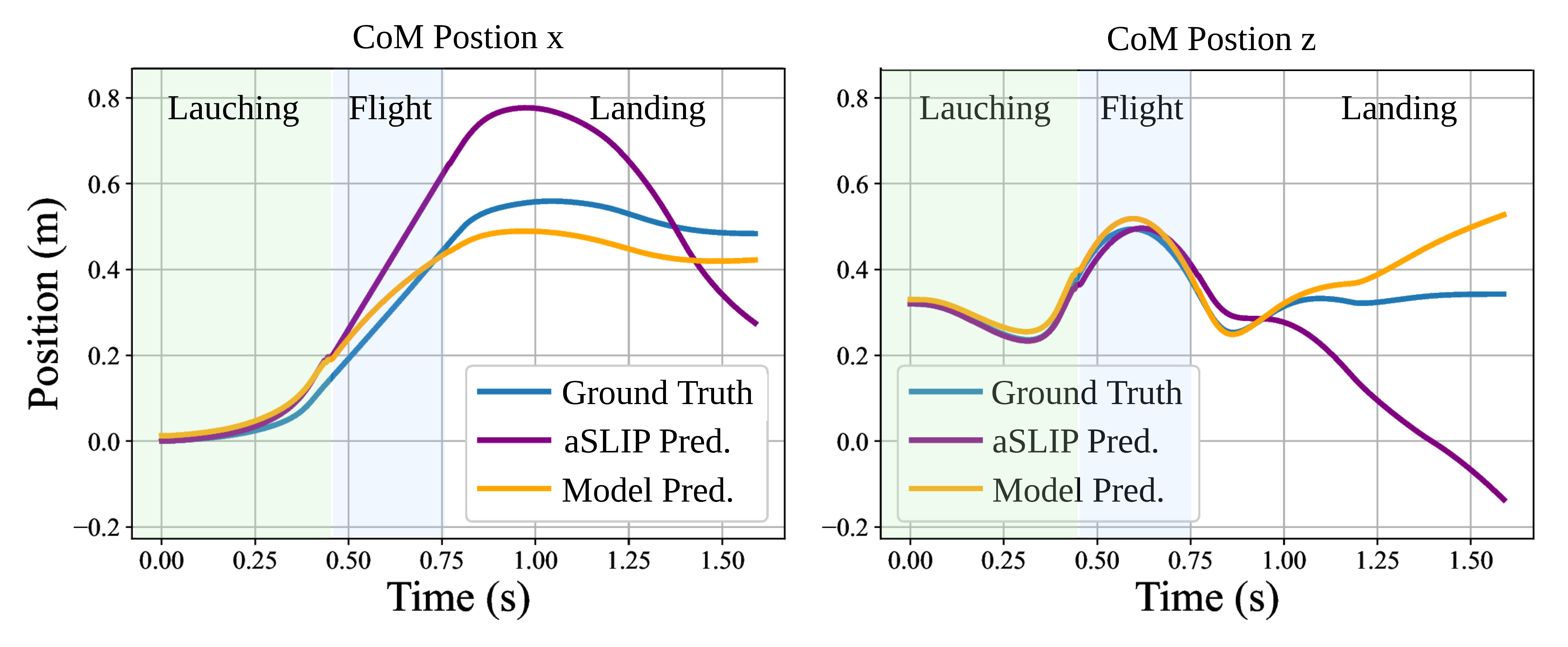}
    \vspace{-8mm}
    \caption{Comparison between the learned model and the aSLIP model regarding the CoM positions prediction, where the Go 1 performs a pronking jumping motion. }
    \label{fig:vs_aslip}
\end{figure}

The prediction results for Go1 pronking are shown in Fig.~\ref{fig:2d_plots}. Table~\ref{tab:result} reports the mean errors and the standard deviation (std) along each direction, demonstrating the high prediction accuracy. To furthermore demonstrate the effectiveness, we regard the actuated SLIP (aSLIP) model (see Appendix.\ref{appendix:reduced_order_model}) as the baseline. Fig.~\ref{fig:vs_aslip} compares the body's linear position along the $x$ and $z$ axes, demonstrating that the learned model outperforms the baseline reduced-order model. Statistical results demonstrate that the average prediction error is reduced from 5.7e-2 m to 1.6e-2 m along the $x$ direction and reduced from 9.5e-2 m to 4.0e-2 m along $z$ direction.
% While the virtual spring in aSLIP effectively models the robot's elasticity during the launch phase, it fails during the landing phase due to the absence of ground interaction. In contrast, the learned model includes ground interaction naturally, allowing trajectory tracking during the landing phase.
 
\subsubsection{Latent Space Dimensionality Tradeoff Analysis}

\begin{figure}
    \centering
    \includegraphics[width=1\linewidth]{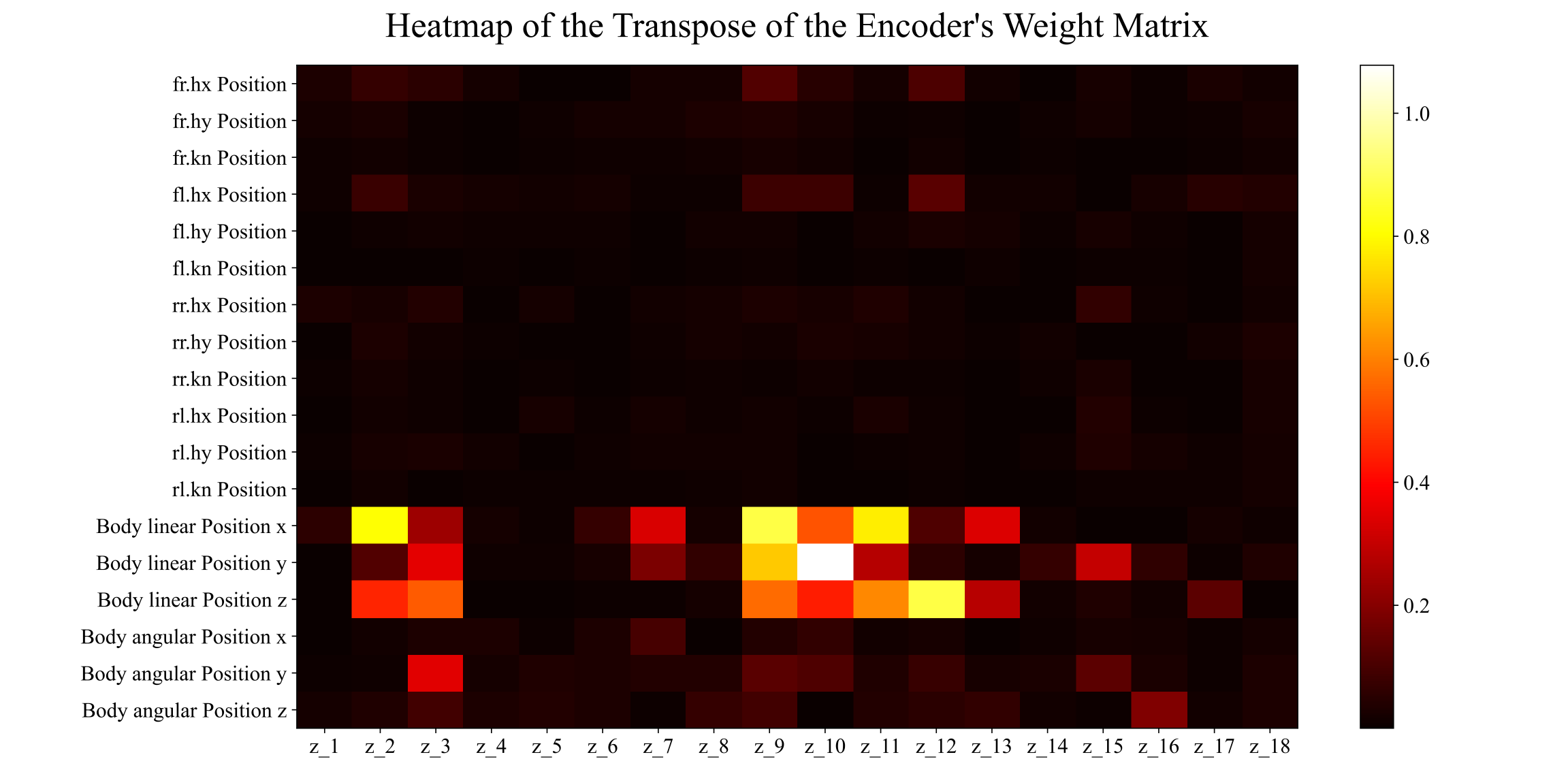}
    \vspace{-7mm}
    \caption{Heatmap of the encoder weight matrix $\mathbf{W}_\mathrm{e}^\mathrm{T}$ (Go1 simulation, 18 latent dimensions).}
    \label{fig:18d_enc}
\end{figure}

\begin{figure}[htb]
    \centering
    \includegraphics[width=1\linewidth]{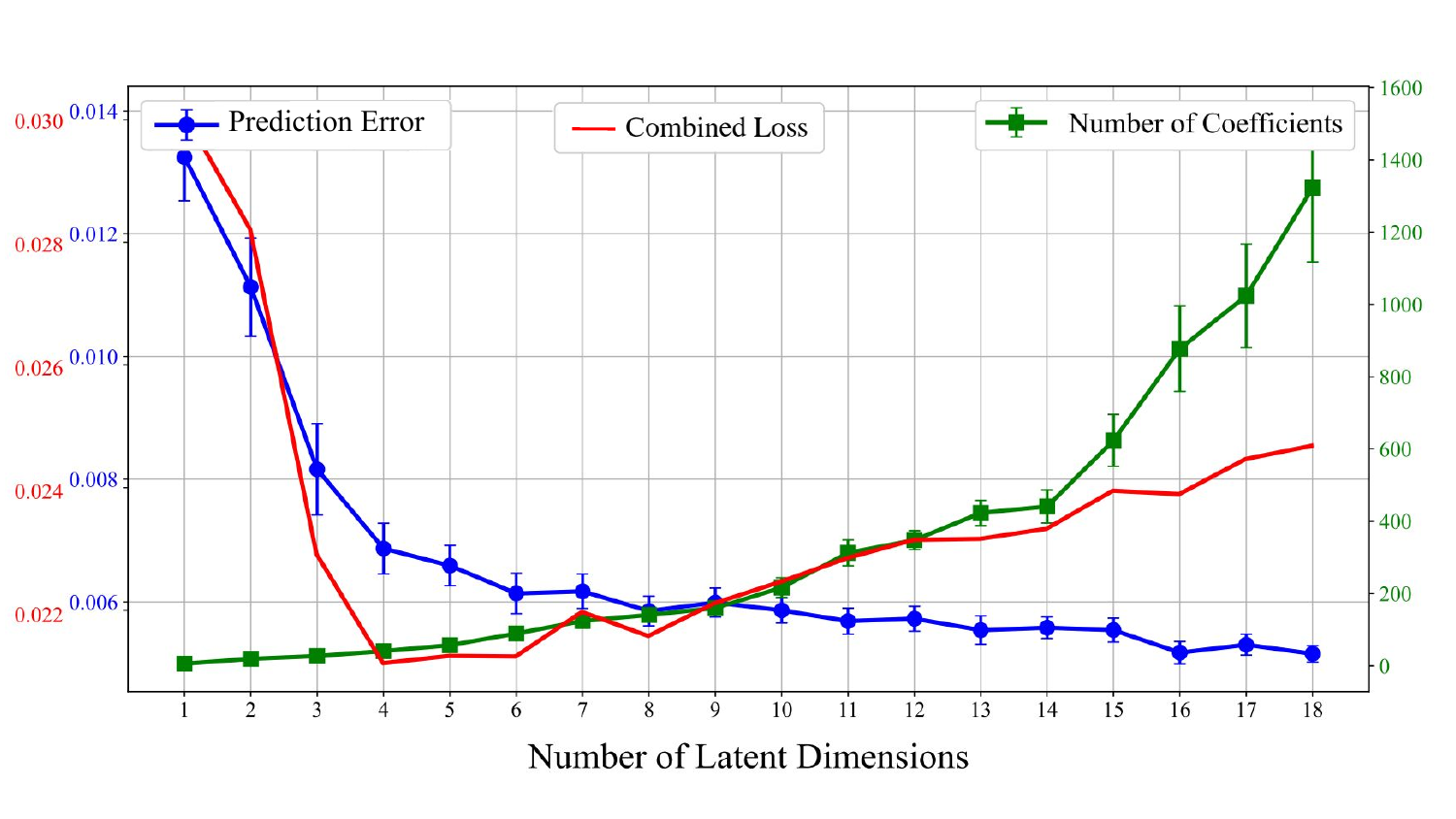}
    \vspace{-8mm}
    \caption{Accuracy \textit{vs.} complexity with the increase of latent space dimensions. %The error bars depict the standard deviation from five different random seeds.
    }
    \label{fig:tradeoff}
\end{figure}

Fig.~\ref{fig:18d_enc} illustrates the effect of latent dimensionality on model design. %It is easy to find that increasing to 18 dimensions does not change the dominance of key full-order components.
Fig.~\ref{fig:tradeoff} further evaluates the trade-off between accuracy (via final reconstruction error $\mathcal{E}_{\text{dec}}$) and complexity (via active coefficients $|\bm{\Xi}|$) across 18 latent dimensions.  
To quantitatively evaluate this trade-off, we employ a combined loss $\mathcal{L}_{\text{mod}}$~\citep{bozdogan1987model}%that is inspired by the Akaike Information Criterion (AIC) for model selection,
, calculated as
\begin{equation}
    \mathcal{L}_{\text{mod}} = 2 \, \mathcal{E}_{\text{dec}} + 2 \, \lambda \,  \log{(|\bm{\Xi}|)},
\end{equation}
with $\lambda = 0.001$.  

The combined loss (red line in Fig.~\ref{fig:tradeoff}) suggests optimal performance with 4–6 latent dimensions. This aligns with the visibly significant components in Fig.~\ref{fig:18d_enc}, where further increasing dimensionality yields marginal gains at the cost of complexity.

\subsection{A1 Simulation Results}
\begin{figure}[t]
  \centering
    % \subfloat[Joint angles]{
    %     % \includegraphics[width=0.48\textwidth]{Images/4d_a1_joints}] }
    %     \includegraphics[width=0.48\textwidth, trim=0 400 0 350, clip]{Images/4d_a1_joints} }
    %     \\
    % \subfloat[CoM positions and orientations]{
        \includegraphics[width=0.48\textwidth]{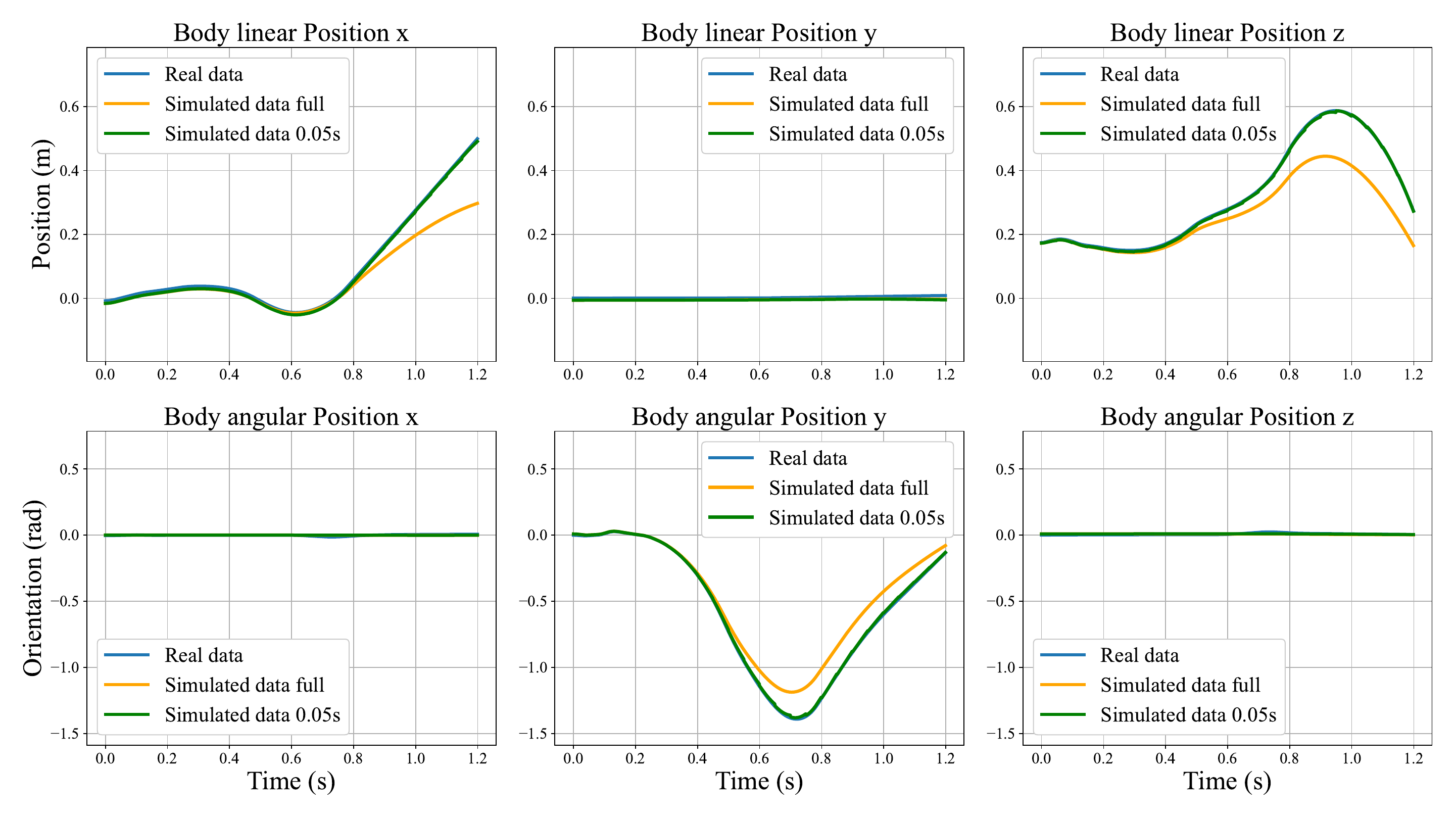}
    % }
    \vspace{-4mm}
  \caption{Learning A1 froggy jumping (simulation data) with four latent dimensions.}
  \label{fig:a1_pos}
\end{figure}

We also conducted experiments with a simulated A1 froggy jump to evaluate the algorithm's effectiveness across different platforms and modalities. The results are illustrated in Fig.~\ref{fig:a1_pos} and Table~\ref{tab:result}.
% \begin{table}[b]
%   \centering
%   \renewcommand{\arraystretch}{1.25}
%   \caption{Prediction performance of the trained model with the simulation A1 data.}
%   \label{tab:result_simulate_A1}  
%   \resizebox{0.49\textwidth}{!}{
%   \begin{tabular}{|c|c|c|c|c|c|c|}
%     \hline  
%     \diagbox[width=7em,dir=NW]{Pre.}{Joint} & fl hx & fl hy & fl hz & rr hx & rr hy & rr hz \\ 
%     \hline
%     Full& 1.0e-10$\pm$1.1e-10 & 0.19$\pm$0.16 & -0.30$\pm$0.34 & 1.0e-10$\pm$1.1e-10 & 0.26$\pm$0.20& -0.32$\pm$0.36 \\ 
%     \hline
%     0.1s& 1.0e-10$\pm$2.6e-10 & 2.8e-3$\pm$3.9e-2 & -3.0e-2$\pm$0.16 & 1.0e-10$\pm$2.6e-10 & 1.4e-2$\pm$5.2e-2& -3.9e-2$\pm$0.23\\ 
%     \hline
%     \diagbox[width=7em,dir=NW]{Pre.}{CoM} & x & y & z & roll & pitch & yaw \\ 
%     \hline
%     Full& -3.5e-2$\pm$5.1e-2 & -8.0e-3$\pm$1.5e-3 & -6.1e-2$\pm$5.8e-2 & 2.4e-3$\pm$4.7e-3& 0.10$\pm$8.3e-2& -8.7e-4$\pm$3.6e-3 \\ 
%     \hline
%     0.1s& -7.9e-3$\pm$1.2e-3 & -8.0e-3$\pm$1.5e-3 & -2.9e-3$\pm$2.0e-3 & 2.4e-3$\pm$4.7e-3& 9.8e-3$\pm$5.6e-2& -8.7e-4$\pm$3.6e-3\\ 
%     \hline    
%   \end{tabular}
%   }
% \end{table}

A notable difference is observed in the body angular position $y$ (i.e., pitch angle) between the Unitree A1 and Go1. The froggy jump causes a more pronounced angular rotation along the $y$-axis in the A1's case. Despite the increased significance of this angular dimension for the froggy jump, the four-dimensional model successfully captures this dynamic and most joint positions. High precision is obtained when reset the latent state every 0.1 s.

\subsection{Go1 Real World Results}
\begin{figure}[t]
  \centering
    \subfloat[Joint angles]{\vspace{-1mm}
        \includegraphics[width=0.48\textwidth, trim=0 400 0 350, clip]{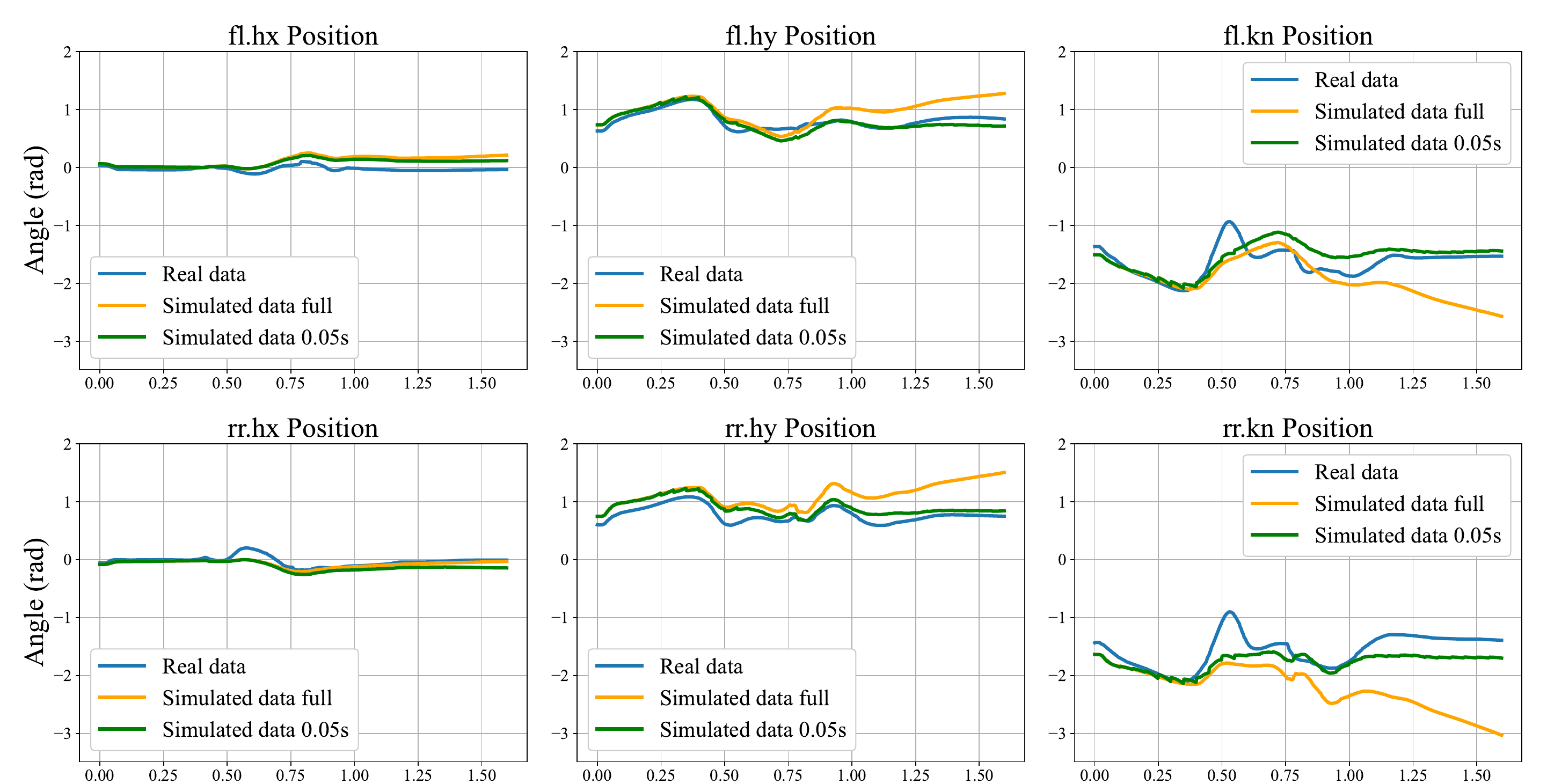}
    }\\
    \subfloat[CoM positions and orientations]{
    \vspace{-2mm}
        \includegraphics[width=0.49\textwidth]{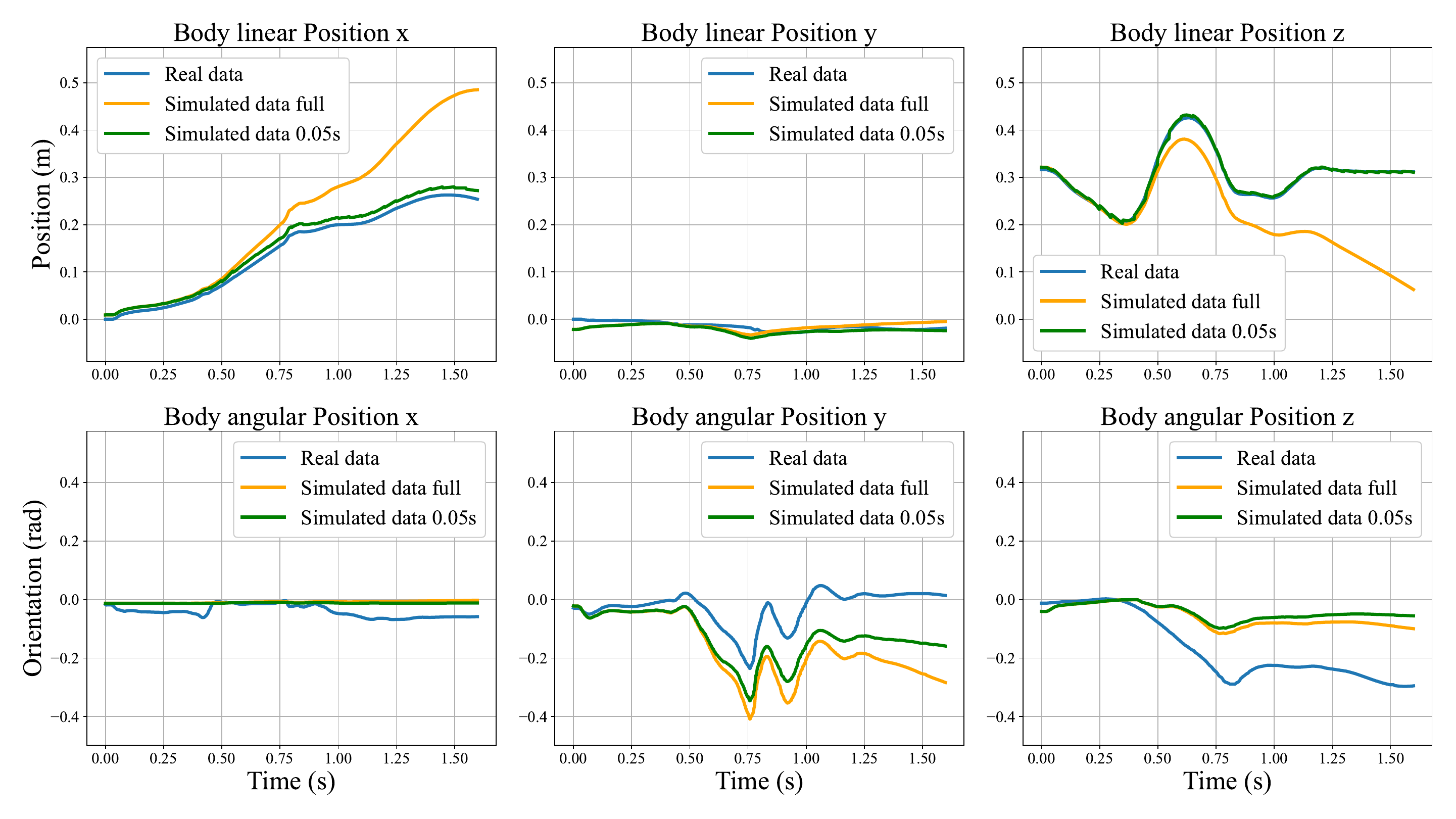}
    }
    \vspace{-2mm}
  \caption{Learning Go1 pronking jumping (real data) with four latent dimensions.}
  \label{fig:real_plots}
\end{figure}

Given the scarcity of real-world compared to simulated data, we first train the model on simulation data with added noise and then subsequently fine-tune it on the few-sample real-world dataset. Fig.~\ref{fig:real_plots} illustrates the model's ability to replicate a jump on the hardware, and Table~\ref{tab:result} reports the prediction error.
% \begin{table}[b]
%   \centering
%   \renewcommand{\arraystretch}{1.25}
%   \caption{Prediction performance of the trained model with the real Go1 data.}
%   \label{tab:result_real_go1}  
%   \resizebox{0.49\textwidth}{!}{
%   \begin{tabular}{|c|c|c|c|c|c|c|}
%     \hline  
%     \diagbox[width=7em,dir=NW]{Pre.}{Joint} & fl hx & fl hy & fl hz & rr hx & rr hy & rr hz \\ 
%     \hline
%     Full& -3.7e-2$\pm$2.6e-2 & 0.22$\pm$0.15 & -0.35$\pm$0.39 & -4.2e-2$\pm$4.7e-2& 0.34$\pm$0.18& -0.65$\pm$0.49 \\ 
%     \hline
%     0.1s& -4.6e-3$\pm$2.3e-2 & 1.7e-2$\pm$8.3e-2 & 3.8e-2$\pm$0.19 &-7.5e-2$\pm$5.1e-2& 0.12$\pm$5.6e-2& -0.21$\pm$0.17\\ 
%     \hline
%     \diagbox[width=7em,dir=NW]{Pre.}{CoM} & x & y & z & roll & pitch & yaw \\ 
%     \hline
%     Full& 7.5e-2$\pm$7.0e-2 & -9.6e-4$\pm$9.6e-3 & -8.1e-2$\pm$7.7e-2 & 3.3e-2$\pm$2.2e-2& -0.14$\pm$9.2e-2& 0.1$\pm$7.8e-2 \\ 
%     \hline
%     0.1s& 1.2e-3$\pm$3.1e-3 & -7.1e-3$\pm$5.4e-3 & 1.2e-3$\pm$3.4e-3 & 2.9e-2$\pm$1.9e-2& -0.10$\pm$5.9e-2& 0.12$\pm$8.9e-2\\ 
%     \hline    
%   \end{tabular}
%   }
% \end{table}

As can be seen from Fig.~\ref{fig:real_plots} and Table~\ref{tab:result}, the proposed model performs well on actual hardware data, despite reduced accuracy during flight and initial landing phases compared to simulations. 
% This discrepancy arises from real system behavior differences, particularly in body angular positions.
The divergence between real and predicted trajectories mainly results from unmodeled actuator and contact effects that accumulate during real execution. Applying a domain randomization in the future work strategy further alleviates this discrepancy by enhancing the model’s robustness to such unmodeled variations. The symbolic model still captures the key motion patterns, with residual errors from friction, delay, and noise remaining bounded.
% Nonetheless, the model effectively manages real-world noise and complexity, demonstrating robust performance.

\section{Discussion and Conclusion}
\label{sec:conclusion}
% The results of our symbolic learning framework show that a reduced latent space can effectively preserve key dynamic patterns of the quadruped robot jumping motion. By combining autoencoder-based embeddings with sparse regression, we obtain interpretable and compact models that approximate the complex full-order dynamics. Our analysis of latent dimensionality shows that increasing beyond 4–6 dimensions offers diminishing returns in accuracy while introducing additional model complexity. The sim-to-real transfer results indicate that the model retains core dynamic features and generalizes well despite limited real-world data, validating its robustness to measurement noise and domain shift.

% In future work, we aim to extend this methodology to more dynamic behaviors, such as walking and running, and to systems involving rapid contact changes or higher-dimensional morphologies, including humanoids.
This work demonstrates that a low-dimensional, physics-aligned latent space can faithfully capture the essential dynamics of quadruped jumping. By combining a linear autoencoder embedding with sparse regression, the proposed framework yields compact and interpretable symbolic models that approximate full-order dynamics with meaningful accuracy. The learned equations recover gravity-like loading, effective damping, and actuation influence, while revealing coupling effects absent in idealized template models yet necessary to represent the richer behaviors of the full quadruped system. Beyond these empirical results, this study highlights how a linear latent embedding—together with a coherent multi-phase training strategy—enables SINDy to discover reduced-order template dynamics for hybrid legged motion, marking a conceptual departure from prior nonlinear SINDy-AE frameworks.

The approach remains sensitive to unmodeled actuator and contact effects, which accumulate during long-horizon rollouts and partly explain the observed sim-to-real discrepancies. Future work will extend the methodology to more diverse locomotion behaviors, incorporate hybrid contact transitions more explicitly, and scale the framework to higher-dimensional morphologies. Another promising direction is to integrate the learned symbolic models into adaptive or phase-dependent controllers for real-time robot movement tasks.

\bibliography{main}             

\appendix
\section{Baseline: actuated SLIP Model}  
\label{appendix:reduced_order_model}
\begin{figure}[htb]
    \centering
    \includegraphics[width=0.3\linewidth]{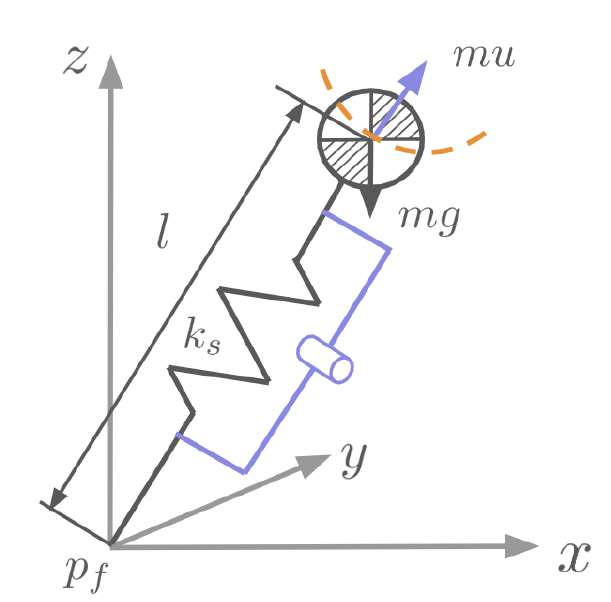}
    \vspace{-3mm}
    \caption{ The aSLIP model. The actuation (denoted as $m\mathbf{u}$) is explicitly captured to enhance controllability.}
    \label{fig:aslip}
\end{figure}

The actuated SLIP (aSLIP)~\citep{ding2024robust} model extends the basic SLIP model by incorporating actuators, allows for enhanced control over legged jumping, as depicted in Fig.~\ref{fig:aslip}. 
The CoM acceleration $\mathbf{\ddot{b}}$ is 
\begin{equation}
    \begin{array}{r l}
        \text{In flight: } &  \mathbf{\ddot{b}} =  \bm{g}\\
        \text{In Contact: }&  \mathbf{\ddot{b}} =   \frac{k_\mathrm{s} |\bm{l}_0 - \bm{l}|\hat{\bm{l}}}{m} + \bm{g} + \bm{u}
    \end{array}
\end{equation}
where $k_\mathrm{s} \in \mathbb{R}^+$ is the spring constant, $m \in \mathbb{R}^+$ is the body mass, $\bm{l}  = \mathbf{b} - \mathbf{p}_k$ is the leg vector, $\hat{\bm{l}}$ is the unit vector
along the leg retraction direction,  $\bm{l}_0$ is the rest length, and $\bm{g} = [0, 0, -g]^\top$ with $g$ being the vertical gravitational constant.
$\bm{u} \in \mathbb{R}^3$ are the driving forces generated by joint motors, which can be seen as the accumulated GRFs on four legs, as calculated using \eqref{eq:grf_torque}.
\end{document}